\documentclass[10pt,twocolumn,letterpaper]{article}

\usepackage{iccv}
\usepackage{times}
\usepackage{epsfig}
\usepackage{graphicx}
\usepackage{amsmath}
\usepackage{amssymb}

\usepackage[breaklinks=true,bookmarks=false]{hyperref}

\iccvfinalcopy 

\def\iccvPaperID{****} 

\ificcvfinal\fi
\begin{document}

\title{UCT: Learning Unified Convolutional Networks for Real-time Visual Tracking}

\author{Zheng Zhu$^{\dagger\ddagger}$\, Guan Huang$^{\S}$, Wei Zou$^{\dagger\ddagger}$\, Dalong Du$^{\S}$, Chang Huang$^{\S}$  \\
$\dagger$ Institute of Automation, Chinese Academy of Sciences\\
$\ddagger$ University of Chinese Academy of Sciences    $\S$ Horizon Robotics, Inc.\\
{\tt\small \{zhuzheng2014,wei.zou\}@ia.ac.cn}   {\tt\small \{guan.huang,dalong.du,chang.huang\}@hobot.cc}
}

\maketitle\thispagestyle{empty}

\begin{abstract}
   Convolutional neural networks (CNN) based tracking approaches have shown favorable performance in recent benchmarks. Nonetheless, the chosen CNN features are always pre-trained in different task and individual components in tracking systems are learned separately, thus the achieved tracking performance may be suboptimal. Besides, most of these trackers are not designed towards real-time applications because of their time-consuming feature extraction and complex optimization details.In this paper, we propose an end-to-end framework to learn the convolutional features and perform the tracking process simultaneously, namely, a unified convolutional tracker (UCT). Specifically, The UCT treats feature extractor and tracking process both as convolution operation and trains them jointly, enabling learned CNN features are tightly coupled to tracking process. In online tracking, an efficient updating method is proposed by introducing peak-versus-noise ratio (PNR) criterion, and scale changes are handled efficiently by incorporating a scale branch into network. The proposed approach results in superior tracking performance, while maintaining real-time speed. The standard UCT and UCT-Lite can track generic objects at 41 FPS and 154 FPS without further optimization, respectively. Experiments are performed on four challenging benchmark tracking datasets: OTB2013, OTB2015, VOT2014 and VOT2015, and our method achieves state-of-the-art results on these benchmarks compared with other real-time trackers.
\end{abstract}

\section{Introduction}

Visual object tracking, which tracks a specified target in a changing video sequence automatically, is a fundamental problem in many aspects such as visual analytics \cite{c5}, automatic driving \cite{c6}, pose estimation \cite{c8} and et al. On the one hand, a core problem of tracking is how to detect and locate the object accurately in the changing scenario such as illumination variations, scale variations, occlusions, shape deformation, and camera motion \cite{c9, c12}. On the other hand, tracking is a time-critical problem because it is always performed in each frame of sequences. Therefore, accuracy, robustness and efficiency are main development directions of the recent tracking approaches.

\begin{figure}[!tp]
  \centering
  \includegraphics[width=1\linewidth]{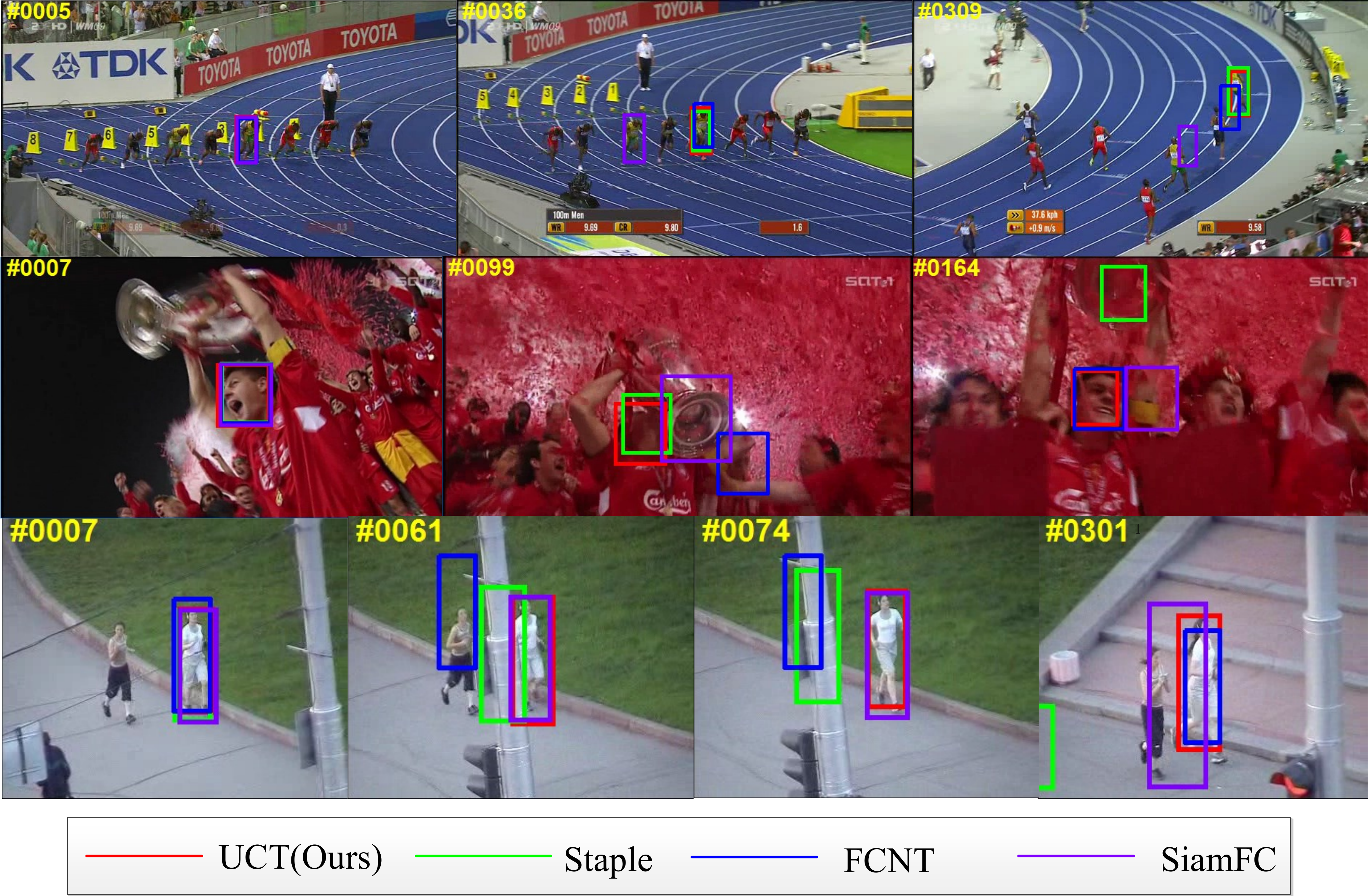}
  \caption{Comparisons of our approach with three state-of-the-art trackers in the changing scenario. The compared trackers are two recent real-time trackers: SiamFC \cite{c33} Staple \cite{c26}, and another fully convolutional tracker FCNT \cite{c36}.}
  \label{figure1}
\end{figure}

As a core components of trackers, appearance model can be divided into generative methods and discriminative methods. In generative model, candidates are searched to minimize reconstruction errors. Representative sparse coding \cite{c4,c7} have been exploited for visual tracking. In discriminative models, tracking is regarded as a classification problem by separating foreground and background. Numerous classifiers have been adapted for object tracking, such as structured support vector machine (SVM) \cite{c2}, boosting \cite{c3} and online multiple instance learning \cite{c1}. Recently, significant attention has been paid to discriminative correlation filters (DCF) based methods \cite{c15, c16, c17, c35} for real-time visual tracking. The DCF trackers can efficiently train a repressor by exploiting the properties of circular correlation and performing the operations in the Fourier domain. Thus conventional DCF trackers can perform at more than 100 FPS \cite{c15, c25}, which is significant for real-time applications.  Many improvements for DCF tracking approaches have also been proposed, such as SAMF \cite{c35} for scale changes, LCT \cite{c17} for long-term tracking, SRDCF \cite{c16} to mitigate boundary effects. The better performance is obtained but the high-speed property of DCF is broken. What is more, all these methods use handcrafted features, which hinder their accuracy and robustness.

Inspired by the success of CNN in object classification \cite{c18, c19}, detection \cite{c20} and segmentation \cite{c21}, the visual tracking community has started to focus on the deep trackers that exploit the strength of CNN in recent years. These deep trackers come from two aspects: one is DCF framework with deep features, which means replacing the handcrafted features with CNN features in DCF trackers \cite{c27, c28}. The other aspect of deep trackers is to design the tracking networks and pre-train them which aim to learn the target-specific features for each new video \cite{c29}. Despite their notable performance, all these approaches separate tracking system into some individual components. What is more,  most of trackers are not designed towards real-time applications because of their time-consuming feature extraction and complex optimization details. For example, the speed of winners in VOT2015 \cite{c10} and VOT2016 \cite{c11} are less than 1 FPS on GPU.

We address these two problems by introducing unified convolutional networks (UCT) to learn the features and perform the tracking process simultaneously. This is an end-to-end and extensible framework for tracking. Specifically, The proposed UCT treats feature extractor and tracking process both as convolution operation, resulting a fully convolutional network architecture. In online tracking, the whole patch can be predicted using the foreground response map by one-pass forward propagation. What is more, efficient model updating and scale handling are proposed to ensure real-time tracking speed.
\subsection{Contributions}
The contributions of this paper can be summarized in three folds as follows:

1, We propose unified convolutional networks to learn the convolutional features and perform the tracking process simultaneously. The feature extractor and tracking process are both treated as convolution operation that can be trained simultaneously. End-to-end training enables learned CNN features are tightly coupled to tracking process.

2, In online tracking, efficient updating and scale handling strategies are incorporated into the tacking framework. The proposed standard UCT (with ResNet-101) and UCT-Lite (with ZF-Net) can track generic objects at 41 FPS and 154 FPS, respectively, which is of significance for real-time computer vision systems.

3, Extensive experiments are carry out on tracking benchmarks and demonstrate that the proposed tracking algorithm performs favorably against existing state-of-the-art methods in terms of accuracy and speed. Figure~\ref{figure1} shows a comparison to state-of-the-art trackers on three benchmark sequences.

\section{Related works}
Visual tracking is a significant problem in computer vision systems and a series of approaches have been successfully proposed for tracking. Since our main contribution is an UCT framework for real-time visual tracking, we give a brief review on three directions closely related to this work: CNN-based trackers,real-time trackers, and fully convolutional networks (FCN).

\subsection{On CNN-based trackers}

Inspired by the success of CNN in object recognition \cite{c18, c19, c20}, researchers in tracking community have started to focus on the deep trackers that exploit the strength of CNN. Since DCF provides an excellent framework for recent tracking research, the first trend is the combination of DCF framework and CNN features. In HCF \cite{c27} and HDT \cite{c28}, the CNN are employed to extract features instead of handcrafted features, and final tracking results are obtained by combining hierarchical response and hedging weak trackers, respectively. DeepSRDCF \cite{c32} exploits shallow CNN features in a spatially regularized DCF framework. Another trend in deep trackers is to design the tracking networks and pre-train them which aim to learn the target-specific features and handle the challenges for each new video. MDNet \cite{c29} trains a small-scale network by multi-domain methods, thus separating domain independent information from domain-specific layers. C-COT \cite{c30} and ECO \cite{c31} employ the implicit interpolation method to solve the learning problem in the continuous spatial domain, where ECO is an improved version of C-COT in performance and speed. These trackers have two major drawbacks: Firstly, they can only tune the hyper-parameters heuristically since feature extraction and tracking process are separate. And they can not end-to-end train and perform tracking systems.  Secondly, none of these trackers are designed towards real-time applications.

\subsection{On real-time trackers}
Other than accuracy and robustness, the speed of the visual tracker is a crucial factor in many real world applications. Therefore, a practical tracking approach should be accurate and robust while operating at real-time. Classical real-time trackers, such as NCC \cite{c22} and Mean-shift \cite{c23}, perform tracking using matching. Recently, discriminative correlation filters (DCF) based methods, which efficiently train a repressor by exploiting the properties of circular correlation and performing the operations in the Fourier domain, have drawn attentions for real-time visual tracking. Conventional DCF trackers such as MOSSE, CSK and KCF can perform at more than 100 FPS \cite{c24, c25, c15}. Subsequently, a series of trackers that follow DCF method are proposed. In DSST algorithm, tracker searches over the scale space for correlation filters to handle the variation of object size. Staple \cite{c26} tracker combines complementary template and color cues in a ridge regression framework. CFLB \cite{c48} and BACF \cite{c49} mitigate the boundary effects of DCF in the Fourier domain. Nevertheless, all these DCF-based trackers employ handcrafted features, which limits better performance.

The recent years have witnessed significant advances of CNN-based real-time tracking approaches. L. Bertinetto et.al \cite{c23} propose a fully convolutional siamese network (SiamFC) to predict motion between two frames. The network is trained off-line and evaluated without any fine-tuning. Similarly to SiamFC, In GOTURN tracker \cite{c34}, the   motion between successive frames is predicted using a deep regression network. These two tackers are able to perform at 86 FPS and 100 FPS respectively on GPU because no fine-tuning is performed. On the one hand, their simplicity and fixed-model nature lead to high speed. On the other hand, this also lose the ability to update the appearance model online which is often critical to account for drastic appearance changes in tracking scenarios. Therefore, there still is an improvement space of performance for real-time deep trackers.

\subsection{On Fully Convolutional trackers}
Fully convolutional networks can efficiently learn to make dense predictions for visual tasks like semantic segmentation, detection as well as tracking. Jonathan Long et al. \cite{c21} transform fully connected layers into convolutional layers to output a heat map for semantic segmentation. The region proposal network (RPN) in Faster R-CNN \cite{c20} is a fully convolutional network that simultaneously predicts object bounds and objectness scores at each position. DenseBox \cite{c37} is an end-to-end FCN detection framework that directly predicts bounding boxes and object class confidences through whole image.
The most related work in tracking literatures is FCNT \cite{c36}, which propose a two-stream fully convolutional network to capture both general and specific object information for visual tracking. However, its tracking components are still independently, so the performance may be impaired. What is more, the FCNT can only perform at 3 FPS on GPU because of its layers switch mechanism and feature map selection method, which hinder it from real-time applications. Compared with FCNT, our UCT treats feature extractor and tracking process in a unified architecture and train them end-to-end, resulting a more compact and much faster tracking approach.

\section{Unified Convolutional networks for tracking}
In this section, the overall architecture of proposed UCT is introduced firstly. Afterwards, we detail the formulation of convolutional operation both in training and test stages.

\begin{figure*}[thpb]
  \centering
  \includegraphics[width=0.8\linewidth]{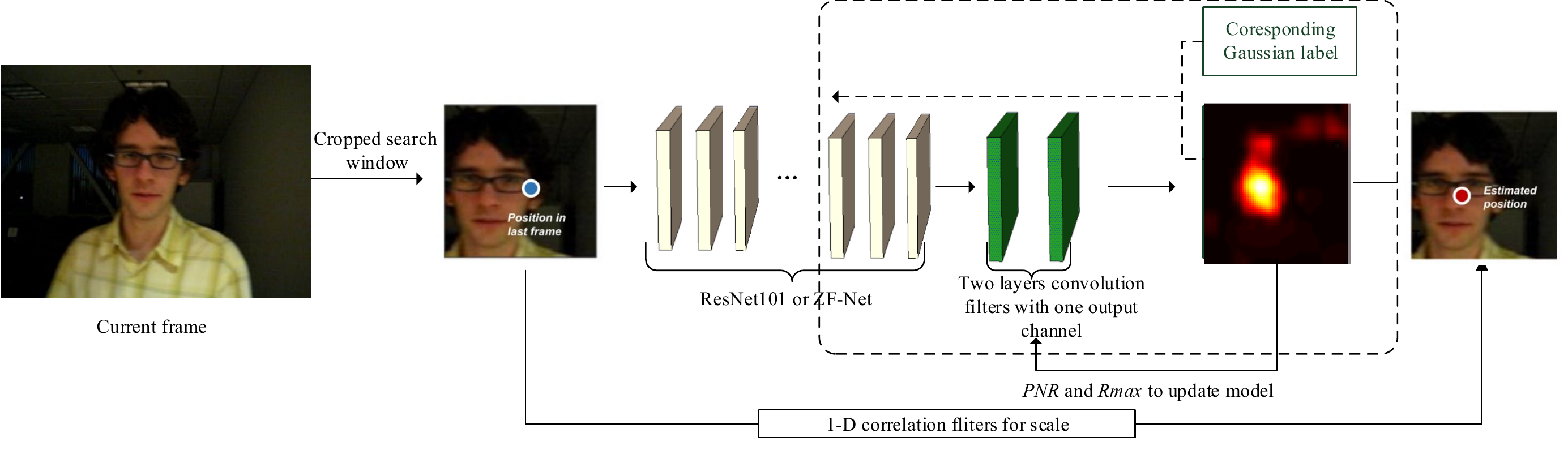}
  \caption{The overall UCT architecture. The solid lines indicate online tracking process, while dashed box and dashed lines indicate off-line training and training on first frame.}
  \label{figure2}
\end{figure*}

\subsection{UCT Architecture}
The overall framework of our tracking approach is a unified convolutional architecture (see Figure~\ref{figure2}), which consists of feature extractor and convolutions performing tracking process. We adopt two groups  convolutional filters to perform tracking process which is trained end-to-end with features extractor. Compared to two-stage approaches adopted in DCF framework within CNN features \cite{c27, c28, c32}, our end-to-end training pipeline is generally preferable. The reason is that the parameters in all components can cooperate to achieve tracking objective.  In Figure ~\ref{figure2}, the search window of current frame is cropped and sent to unified convolutional networks. The estimated new target position is obtained by finding the maximum value of the response map. Another separate 1-dimentioanl convolutional branch is used to estimate target scale and model updating is performed if necessary.  The solid lines indicate online tracking process, while dashed box and dashed lines are included in off-line training and training on first frame. Each feature channel in the extracted sample is always multiplied by a Hann window, as described in \cite{c15}.

\subsection{Formulation}
In the UCT formulation, the aim is to learn a series of convolution filters $f$ from training samples ${(x_k, y_k)}_{k=1:t}$. Each sample is extracted using another CNN from an image region. Assuming sample has the spatial size $M \times N$, the output has the spatial size $m \times n$ ($m=M / stride_M, n=N / stride_N$). The desired output $y_k$ is a response map which includes a target score for each location in the sample $x_k$. The convolutional response of the filter on sample $x$ is given by
\begin{equation}
\label{eq1}
R(x) = \sum_{l=1}^dx^l*f^l
\end{equation}
where $x^l$ and $f^l$ is $l$-th channel of extracted CNN features and desired filters, respectively, $*$ denotes convolutional operation. The filter can be trained by minimizing $L_2$ loss which is obtained between the response $R(x_k)$ on sample $x_k$ and the corresponding Gaussian label $y_k$
\begin{equation}
\label{eq2}
L = {||R(x_k) - y_k||}^2 + \lambda\sum_{l=1}^d{||f^l||}^2
\end{equation}
The second term in~(\ref{eq2}) is a regularization with a weight parameter $\lambda$.

In test stage, the trained filters are used to evaluate an image patch centered around the predicted target location. The evaluation is applied in a sliding-window manner, thus can be operated as convolution:
\begin{equation}
\label{eq3}
R(z) = \sum_{l=1}^dz^l*f^l
\end{equation}
Where $z$ denote the feature map extracted from last target position including context.

It is noticed that the formulation in our framework is similar to DCF, which solve this ridge regression problem in frequency domain by circularly shifting the sample. Different from DCF, we adopt gradient descent to solve equation~(\ref{eq2}), resulting in convolution operations. Noting that the sample $x_k$ is also extracted by CNN, these convolution operations can be naturally unified in a fully convolutional network. Compared to DCF framework, our approach has three advantages: firstly, both feature extraction and tracking convolutions can be pre-trained simultaneously, while DCF based trackers can only tune the hyper-parameters heuristically. Secondly, model updating can be performed by SGD, which maintains the long-term memory of target appearance. Lastly, our framework is much faster than DCF framework within CNN features.
\subsection{Training}
Since the objective function defined in equation~(\ref{eq2}) is convex, it is possible to obtain the approximate global optima via gradient descent with an appropriate learning rate in limited steps. We divide the training process into two periods: off-line training that can encode the prior tracking knowledge, and the training on first frame to adapt to specific target.

In off-line training, the goal is to minimize the loss function in equation~(\ref{eq2}). In tracking, the target position in last frame is always not centered in current cropped patch. So for each image, the train patch centered at the given object is cropped with jittering. The jittering consists of translation and scale jittering, which approximates the variation in adjacent frames when tracking.   Above cropped patch also includes background information as context. In training, the final response map is obtained by last convolution layer within one channel. The label is generated using a Gaussian function with variances proportional to the width and height of object. Then the $L_2$ loss can be generated and the gradient descent can be performed to minimize equation ~(\ref{eq2}). In this stage, the overall network consists of a pre-trained network with ImageNet (ResNet101 in UCT and ZF-Net in UCT-Lite) and following convolutional filters. Last part of ResNet or ZF-Net is trained to encode the prior tracking knowledge with following convolutional filters, making the extracted feature more suitable for tracking.

The goal of training on first frame is to adapt to a specific target. The network architecture follows that in off-line training, while later convolutional filters are randomly initialized by zero-mean Gaussian distribution. Only these randomly initialized layers are trained using SGD in first frame.

\begin{figure}[thpb]
  \centering
  \includegraphics[width=1\linewidth]{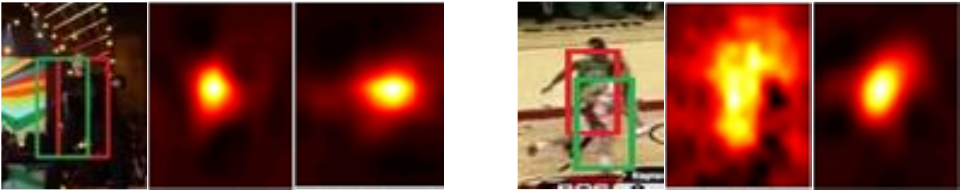}
  \caption{From left to right: images, response maps without off-line training and response maps with off-line training. Green and red boxes in images indicates tracking results without and with off-line training, respectively.}
  \label{figure3}
\end{figure}

Off-line training encodes prior tracking knowledge and constitute a tailored feature extractor. We perform online tracking with and without off-line training to illustrate this effect. In Figure ~\ref{figure3}, we show tracking results and corresponding response maps without or with off-line training. In left part of Figure~\ref{figure3}, the target singer is moving to right, the response map with off-line training effectively reflects this translation changes while response map without off-line training are not capable of doing this. So the tracker without off-line training misses this critical frame. In right part of Figure~\ref{figure3}, the target player is occluded by another player, the response map without off-line training becomes fluctuated and tracking result is effected by distractor, while response map with off-line training still keeps discriminative. The results are somewhat unsurprising, since CNN features trained on ImageNet classification data are expected to have greater invariance to position and same class. In contrast, we can obtain more suitable feature tracking by end-to-end off-line training.

\section{Online tracking}

After off-line training and training on the first frame, the learned network is used to perform online tracking by equation~(\ref{eq3}). The estimate of the current target state is obtained by finding the maximum response score. Since we use a fully convolutional network architecture to perform tracking, the whole patch can be predicted using the foreground heat map by one-pass forward propagation. Redundant computation was saved. Whereas in \cite{c29} and \cite{c38}, network has to be evaluated for $N$ times given $N$ samples cropped from the frame. The overlap between patches leads to a lot of redundant computation.

\subsection{Model update}
Most of tracking approaches update their model in each frame or at a fixed interval \cite{c15, c25, c27, c30, c31}. However, this strategy may introduce false background information when the tracking is inaccurate, target is occluded or out of view.  In the proposed method, model update is decided by evaluating the tracking results. Specifically, we consider the maximum value in the response map and the distribution of other response value simultaneously.

\begin{figure}[thpb]
  \centering
  \includegraphics[width=1\linewidth]{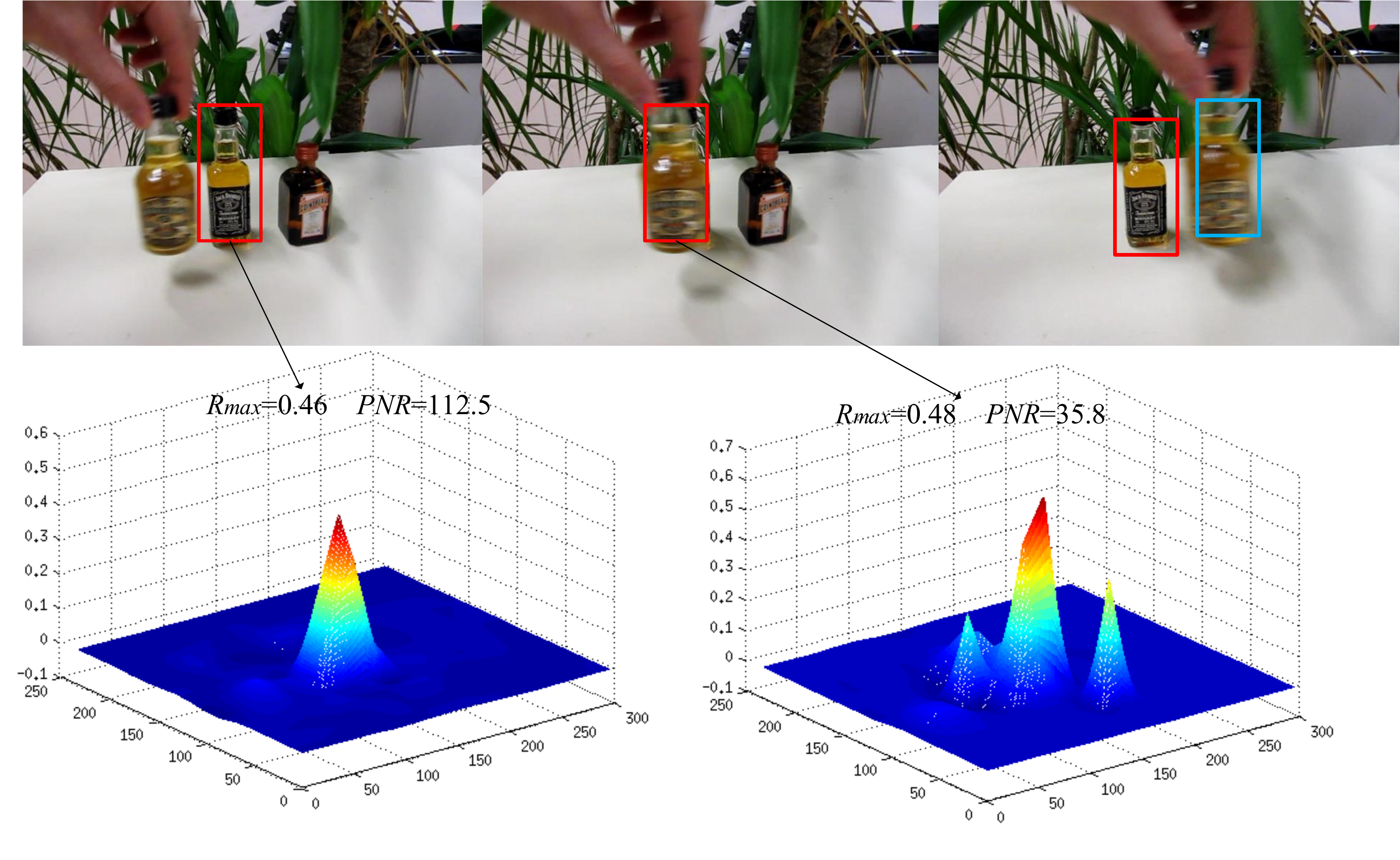}
  \caption{Updating results of UCT and UCT$\_$No$\_$\emph{PNR} (UCT without \emph{PNR} criterion). The first row shows frames that the target is occluded by distractor. The second row are corresponding response maps. $R_{max}$ still keeps large in occlusion while \emph{PNR} significantly decreases. So the unwanted updating is avoided by considering \emph{PNR} constraint simultaneously. The red and blue boxes in last image are tracking results of UCT and UCT$\_$No$\_$\emph{PNR}, respectively.}
  \label{figure4}
\end{figure}

Ideal response map should have only one peak value in actual target position and the other values are small.  On the contrary, the response will fluctuate intensely and include more peak values as shown in Figure~\ref{figure4}. We introduce a novel criterion called peak-versus-noise ratio (\emph{PNR}) to reveal the distribution of response map. The \emph{PNR} is defined as

\begin{equation}
\label{eq4}
\emph{PNR} = \frac{R_{max}-R_{min}}{mean(R{\backslash}R_{max})}
\end{equation} where

\begin{equation}
R_{max} = \max{R(z)}
\end{equation} and $R_{min}$ is corresponding minimum value of response map. Denominator in equation~(\ref{eq4}) represents mean value of response map except maximum value and is used to measure the noise approximately. The \emph{PNR} criterion becomes larger when response map has fewer noise and sharper peak. Otherwise, the \emph{PNR} criterion will fall into a smaller value. We save the \emph{PNR} and $R_{max}$ and calculate their historical average values as threshold:

\begin{equation}
\label{eq6}
\left\{
     \begin{array}{rl}
     \emph{PNR}_{threshold} =& \frac{\sum_{t=1}^T{PNR}_t}{T}  \\
     R_{threshold} =& \frac{\sum_{t=1}^T{R}_{max}^t}{T}
     \end{array}
\right.
\end{equation} Model update is performed only when both two criterions in equation~(\ref{eq6}) are satisfied. The updating is one step SGD with smaller learning rate compared with that in the first frame. Figure~\ref{figure4} illustrates the necessity of proposed \emph{PNR} criterion by showing tracking results under occlusions. As shown in Figure~\ref{figure4}, updating is still performed if only according to $R_{max}$ criterion when target is under occlusion. Introduced noise will result in inaccurate tracking results even failures. The \emph{PNR} criterion significantly decreases in these unreliable frames thus avoids unwanted updating.

\subsection{Scale estimation}

A conventional approach of incorporating scale estimation is to evaluate the appearance model at multiple resolutions by performing an exhaustive scale search \cite{c35}. However, this search strategy is computationally demanding and not suitable for real-time tracking. Inspired by \cite{c45}, we introduce a 1-dimensional convolutional filters branch to estimate the target size as shown in Figure~\ref{figure2}.  This scale filter is applied at an image location to compute response scores in the scale dimension, whose maximum value can be used to estimate the target scale. Such learning separate convolutional filters to explicitly handle the scale changes is more efficient for real-time tracking.

In training and updating of scale convolutional filters, the sample $x$ is extracted from variable patch sizes centered around the target:

\begin{equation}
\label{eq7}
size(P^n) = a^nW \times a^nH~~~n \in \{-\lfloor\frac{S-1}{2}\rfloor,...,\lfloor\frac{S-1}{2}\rfloor\}
\end{equation}
Where $S$ is the size of scale convolutional filters, $W$ and $H$ are the current target size, $a$ is the scale factor. In scale estimation test, the sample is extracted using the same way after translation filters are performed. Then the scale changes compared to previous frame can be obtained by maximizing the response score. Note that the scale estimation is performed only when model updating condition is satisfied.

\section{Experiments}
Experiments are performed on four challenging tracking datasets: OTB2013 with 50 videos, OTB2015 with 100 videos, VOT2014 with 25 videos and VOT2015 with 60 videos . All the tracking results are using the reported results to ensure a fair comparison.

\subsection{Implement details}

We adopt ResNet-101 in standard UCT and ZF-Net in UCT-Lite as feature extractor, respectively. In off-line training, last four layers of ResNet and last two layers of ZF-Net are trained. Our training data comes from UAV123 \cite{c47}, and TC128 \cite{c13} excluding the videos that overlap with test set. In each frame, patch is cropped around ground truth and resized into 224*224. The translation and scale jittering are 0.05 and 0.03, respectively.  We apply stochastic gradient descent (SGD) with momentum of 0.9 to train the network and set the weight decay $\lambda$ to 0.005. The model is trained for 30 epochs with a learning rate of $10^{-5}$. In online training on first frame, SGD is performed 50 steps with a learning rate of $5*10^{-7}$ and $\lambda$ is set to 0.01. In online tracking, the model update is performed by one step SGD with a learning rate of $1*10^{-7}$. S and a in equation~(\ref{eq7}) is set to 33 and 1.02, respectively.

The proposed UCT is implemented using Caffe \cite{c39} with Matlab wrapper on a PC with an Intel i7 6700 CPU, 48 GB RAM, Nvidia GTX TITAN X GPU. The code and results will be made publicly available.

\subsection{Results on OTB2013}
\label{resultsonOTB2013}
OTB2013 \cite{c14} contains 50 fully annotated sequences that are collected from commonly used tracking sequences. The evaluation is based on two metrics: precision plot and success plot. The precision plot shows the percentage of frames that the tracking results are within certain distance determined by given threshold to the ground truth. The value when threshold is 20 pixels is always taken as the representative precision score. The success plot shows the ratios of successful frames when the threshold varies from 0 to 1, where a successful frame means its overlap is larger than this given threshold. The area under curve (AUC) of each success plot is used to rank the tracking algorithm.

In this experiment, ablation analyses are performed to illustrate the effectiveness of proposed component at first. Then we compare our method against the three best trackers that presented in the OTB2013, Struck \cite{c2}, SCM \cite{c42} and TLD \cite{c43}. We also include recent real-time trackers presented at top conferences and journals, they are KCF (T-PAMI 2015) \cite{c15}, Siamese-FC (ECCV 2016) \cite{c33}, Staple (CVPR 2016) \cite{c26}, SCT (CVPR 2016) \cite{c34}. What is more, other recent trackers, HDT (CVPR2016) \cite{c28}, FCNT (ICCV 2015) \cite{c36}, CNN-SVM (ICML 2015) \cite{c40}, DLSSVM (CVPR2016) \cite{c41} and HCF (ICCV2015) \cite{c27} are also compared, these approaches are not real-time but most of their speed is more than 10FPS. There are five deep trackers and seven shallow trackers in total. The one-pass evaluation (OPE) is employed to compare these trackers.

\begin{figure}[thpb]
\centering
\includegraphics[width=1.0\linewidth]{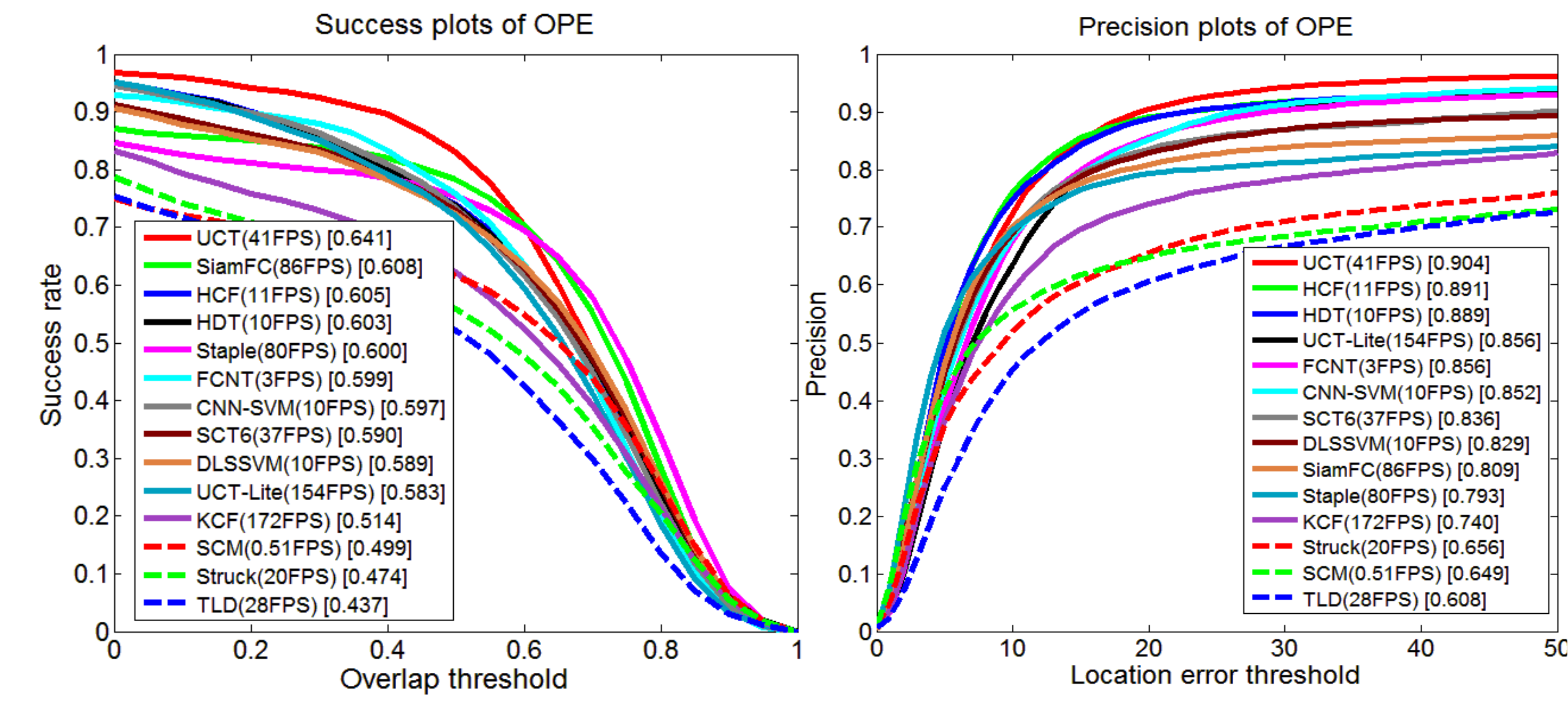}

\caption{ Precision and success plots on OTB2013 \cite{c14}. The numbers in the legend indicate the representative precisions at 20 pixels for precision plots, and the area-under-curve scores for success plots.}
\label{figure5}
\end{figure}

\subsubsection{Ablation analyses}
To verify the contribution of each component in our algorithm, we implement and evaluate four variations of our approach: Firstly, the effectiveness of our off-line training is tested by comparison without this procedure (UCT$\_$No$\_$Off-line), where the network is only trained within the first frame of a specific sequence. Secondly, the tracking algorithm that updates model without \emph{PNR} constraint (UCT$\_$No$\_$\emph{PNR}, only depends on $R_{max}$) is compared with the proposed efficient updating method. Last two additional versions are UCT within multi-resolutions scale (UCT$\_$MulRes$\_$Scale) and without scale handling (UCT$\_$No$\_$Scale).

As shown in Table 1, the performances of all the variations are not as good as our full algorithm (UCT) and each component in our tracking algorithm is helpful to improve performance. Specifically, Off-line training encodes prior tracking knowledge and constitute a tailored feature extractor, so the UCT outperforms UCT$\_$No$\_$Off-line with a large margin. Proposed \emph{PNR} constraint for model update improves performance as well as speed, since it avoids updating in unreliable frames. Although exhaustive scale method in multiple resolutions improves the performance of tracker, it brings higher computational cost. By contrast, learning separate filters for scale in our approach gets a better performance while being computationally efficient.

\begin{table}[t]
  \centering
 \begin{tabular}{cccc}
    \hline
    \bf Approaches     & \bf AUC     & \bf Precision20 & \bf Speed (FPS) \\
    \hline
    UCT$\_$No$\_$Off-line & 0.601  & 0.863 &  41     \\
    UCT$\_$No$\_$\emph{PNR}      & 0.624  & 0.880 &  33      \\
    UCT$\_$No$\_$Scale    & 0.613  & 0.871 &  51     \\
    UCT$\_$MulRes$\_$Scale& 0.629  & 0.893 &  22     \\
    UCT                   & 0.641  & 0.904 &  41     \\
    \hline

  \end{tabular}

  \centerline {\caption{ Performance on OTB2013 of UCT and its variations}}
   \label{table1}
\end{table}

\subsubsection{Comparison with state-of-the-art trackers}
We compare our method against the state-of-the-art trackers as shown in~\ref{resultsonOTB2013}. There are five deep trackers and seven shallow trackers in total. Figure~\ref{figure5} illustrates the precision and success plots based on center location error and bounding box overlap ratio, respectively. It clearly illustrates that our algorithm, denoted by UCT, outperforms the state-of-the-art trackers significantly in both measures. In success plot, our approach obtain an AUC score of 0.641, significantly outperforms SiamFC  and HCF by 3.3\% and 3.6\%, respectively.  In precision plot, our approach obtains a score of 0.904, outperforms HCF and HDT by 1.3\% and 1.5\%, respectively. It worth mentioning that our UCT provides significantly better performance while being 13 times faster compared to the FCNT tracker.

The top performance can be attributed to that our methods encodes prior tracking knowledge by off-line training and extracted features is more suitable for following tracking convolution operations. By contrast, the CNN features in other trackers are always pre-trained in different task and is independently with the tracking process, thus the achieved tracking performance may not be optimal. What is more, efficient updating and scale handling strategies ensure robustness and speed of the tracker.

Besides standard UCT, we also implement a lite version of UCT (UCT-Lite) which adopts ZF-Net \cite{c46} and ignores scale changes. As shown in Figure~\ref{figure5}, the UCT-Lite obtains a precision score of 0.856 while operates at 154 FPS. Our UCT-Lite approach is much faster than recent real-time trackers, SiamFC and Staple, while significantly outperforms them in precision.

\begin{figure}[thpb]
\centering
\includegraphics[width=1.0\linewidth]{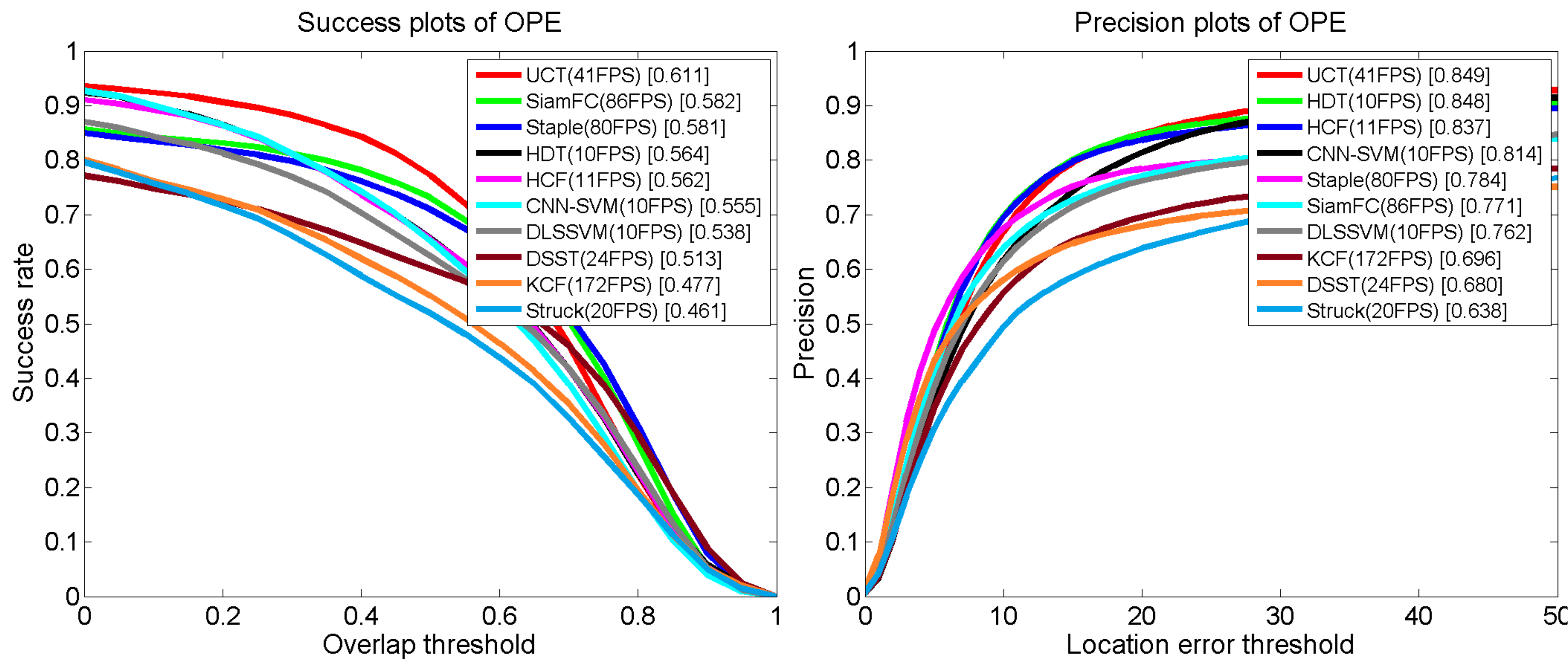}
\caption{ Precision and success plots on OTB2015 \cite{c9}. The numbers in the legend indicate the representative precisions at 20 pixels for precision plots, and the area-under-curve scores for success plots.}
\label{figure6}
\end{figure}

\begin{figure*}[thpb]
  \centering
  \includegraphics[width=1\linewidth]{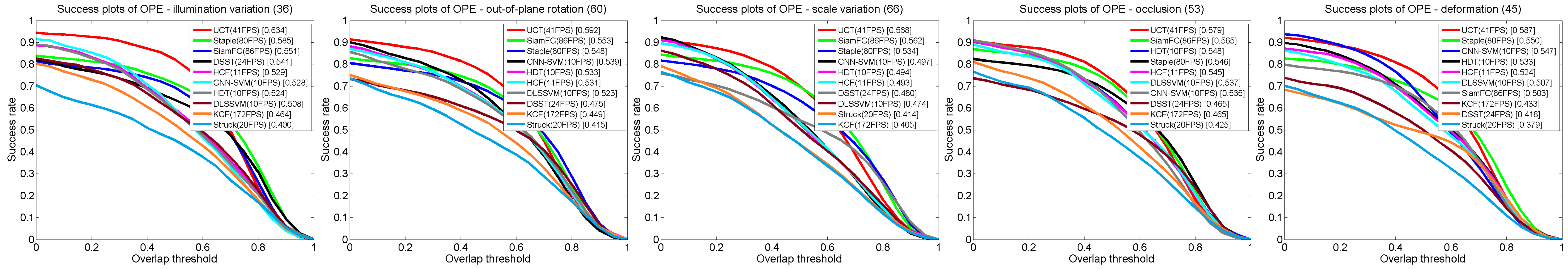}
  \caption{The success plots of OTB2015 \cite{c9} for five challenge attributes: illumination variation, out-of-plane rotation, scale variation, occlusion deformation and background clutter. In the caption of each sub-figure, the number in parentheses denotes the number of the video sequences in the corresponding situation.}
  \label{figure7}
\end{figure*}

\subsection{Results on OTB2015}
OTB2015 \cite{c9} is the extension of OTB2013 and contains 100 video sequences. Some new sequences are more difficult to track. In this experiment, we compare our method against the best trackers that presented in the OTB2015, Struck \cite{c2}. What is more, some recent trackers are also compared, they are KCF (T-PAMI 2015) \cite{c15}, DSST (T-PAMI 2017) \cite{c45}, SiamFC (ECCV 2016) \cite{c33}, Staple (CVPR 2016) \cite{c26}, HDT (CVPR2016) \cite{c28}, HCF (ICCV2015) \cite{c27}, FCNT (ICCV 2015) \cite{c36}, DLSSVM (CVPR2016) \cite{c41} and CNN-SVM (ICML 2015) \cite{c40}. There are five deep trackers and four shallow trackers in total. The one-pass evaluation (OPE) is employed to compare these trackers.

Figure~\ref{figure6} illustrates the precision and success plots of compared trackers, respectively. The proposed UCT approach outperforms all the other trackers in terms of both precision score and success score. Specifically, our method achieves a success score of 0.611, which outperforms the SiamFC (0.582) and Staple (0.581) method with a large margin. Since the proposed tracker adopts a unified convolutional architecture and efficient online tracking strategies, it achieves superior tracking performance and real-time speed.

For detailed performance analysis, we also report the results on various challenge attributes in OTB2015, such as illumination variation, scale changes, occlusion, etc. Figure~\ref{figure7} demonstrates that our tracker effectively handles these challenging situations while other trackers obtain lower scores. Comparisons of our approach with three state-of-the-art trackers in the changing scenario is shown in Figure~\ref{figure1}.

\subsection{Results on VOT}

The Visual Object Tracking (VOT) challenges are well-known competitions in tracking community. The VOT have held several times from 2013 and their results will be reported at ICCV or ECCV. In this subsection, we compare our method, UCT with entries in VOT 2014 \cite{c44} and VOT2015 \cite{c10}.

VOT2014 contains 25 sequences with substantial variations. A tracker is re-initialized whenever tracking fails and the evaluation module reports both accuracy and robustness, which correspond to the bounding box overlap ratio and the number of failures, respectively. There are two sets of experiments: trackers are initialized with either ground-truth bounding boxes (baseline) or randomly perturbed ones (region noise). The VOT evaluation then provides a ranking analysis based on both statistical and practical significance of the performance gap between trackers. We compare our algorithm with the top 7 trackers in VOT2014 challenges \cite{c44}. What is more, we add additional three state-of-the-art real-time trackers GOTURN (ECCV2016) \cite{c34}, SiamFC (ECCV2016 Workshop) \cite{c33} and Staple (CVPR2016) \cite{c26}.

\begin{figure}[thpb]
\centering
\includegraphics[width=0.9\linewidth]{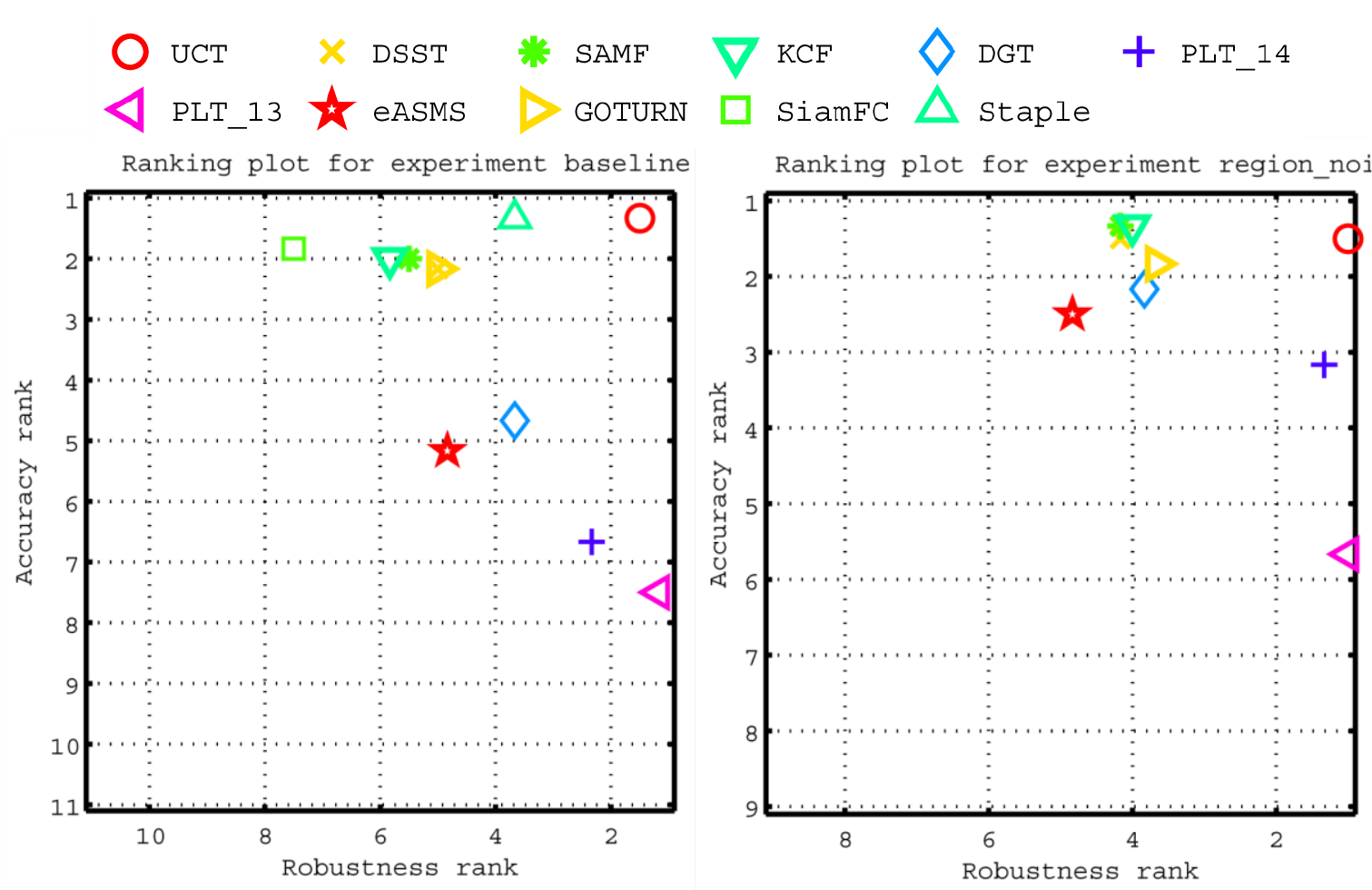}
\caption{Accuracy and robustness rank plot on VOT2014.The better trackers are located at the upper-right corner.}
\label{figure8}
\end{figure}

As shown in Figure~\ref{figure8}, proposed UCT is ranked top both in accuracy and robustness. With precise re-initializations, UCT ranks second both in accuracy and robustness while comprehensive performance is best. It worth mentioning that UCT significantly outperforms three state-of-the-art real-time trackers in robustness rank. The similar performance is obtained with imprecise re- initializations as shown in region noise experiment results, which implies that out UCT can achieve long-term tracking within a re-detection module.

VOT2015 \cite{c10} consists of 60 challenging videos that are automatically selected from a 356 sequences pool. The trackers in VOT2015 is evaluated by expected average overlap (EAO) measure, which is the inner product of the empirically estimating the average overlap and the typical-sequence-length distribution. The EAO measures the expected no-reset overlap of a tracker run on a short-term sequence.

\begin{figure}[thpb]
\centering
\includegraphics[width=1\linewidth]{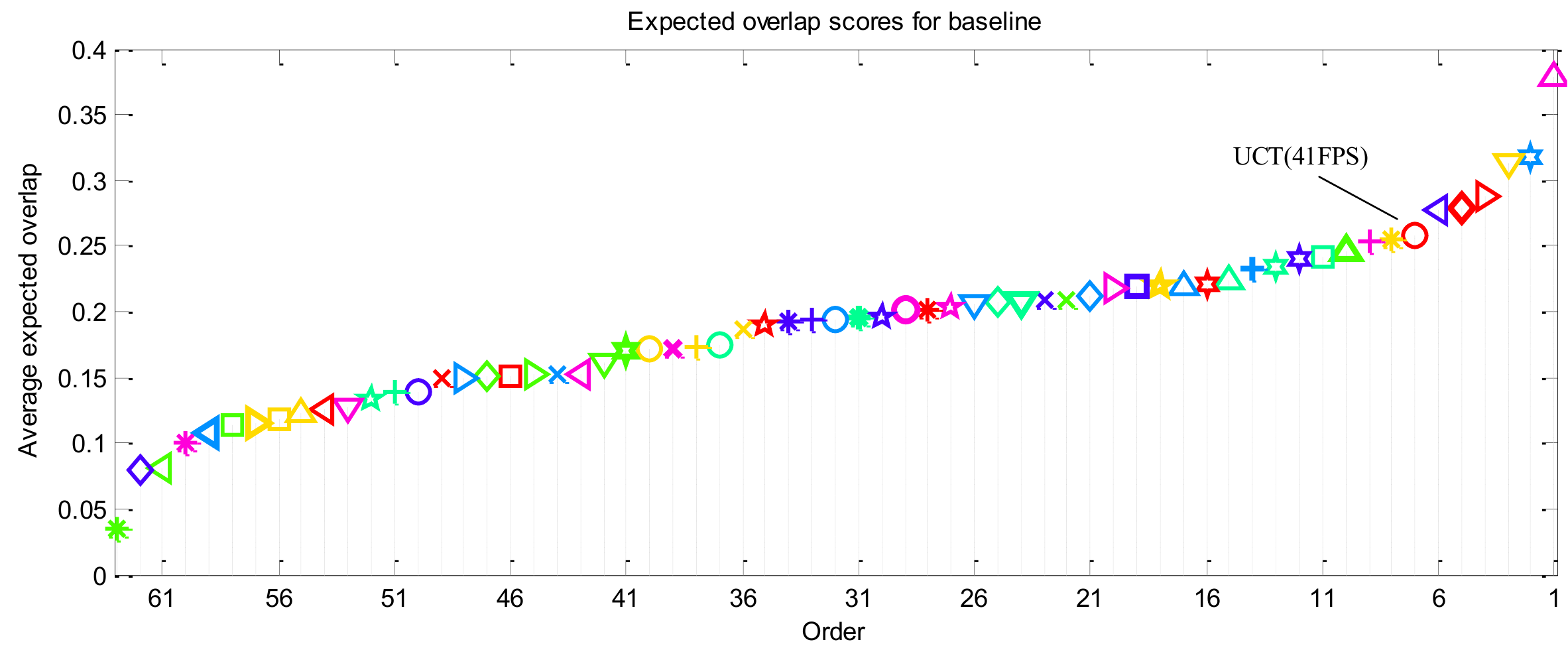}
\caption{EAO rank plot on VOT2015. The better trackers are located at the right. The ranking of other trackers is consistent with VOT2015.}
\label{figure9}
\end{figure}

 Figure~\ref{figure9} illustrates that proposed UCT can ranked seventh in EAO measures. None of top six trackers can perform in real-time(their speed is less than 5 EFO). Since UCT employs end-to-end training, efficient updating and scale handling strategies, it can achieve a great balance between performance and speed.

\section{Conclusions}
In this work, we proposed a unified convolutional tracker (UCT) that learn the convolutional features and perform the tracking process simultaneously. In online tracking, efficient updating and scale handling strategies are incorporated into the network. It is worth to emphasize that our proposed algorithm not only performs superiorly, but also runs at a very fast speed which is significant for real-time applications. Experiments are performed OTB2013, OTB2015, VOT2014 and VOT2015, and our method achieves state-of-the-art results on these benchmarks compared with other real-time trackers.

\section*{Acknowledgment}
This work was done when Zheng Zhu was an intern at Horizon Robotics, Inc. This work is supported in part by the National Natural Science Foundation of China under Grant No. 61403378 and 51405484, and in part by the National High Technology Research and Development Program of China under Grant No.2015AA042307.

{\small
\bibliographystyle{ieee}

\end{document}